%% file: main.tex
\documentclass[letterpaper, 10 pt, conference]{ieeeconf}  

\IEEEoverridecommandlockouts                              

\usepackage{amsmath}
\usepackage{amssymb}
\usepackage{amsfonts}
\usepackage{bbm}
\usepackage{booktabs, multirow} 
\usepackage{soul}
\usepackage{changepage,threeparttable} 
\usepackage[table]{xcolor} 
\usepackage{makecell}
\usepackage[ruled,vlined]{algorithm2e}
\usepackage{mdframed}
\usepackage{graphicx}
\usepackage[noadjust]{cite}

\title{\LARGE \bf
        Manipulating Neural Path Planners via Slight Perturbations 
}

\author{Zikang Xiong and Suresh Jagannathan 
\thanks{ Authors are with the Computer Science Department, Purdue University, 
        West Lafayette, IN, USA. {\tt\small \{xiong84,suresh\}@cs.purdue.edu}
}
}

\begin{document}

\thispagestyle{empty}
\pagestyle{empty}
\input{macros.tex}

\maketitle

\begin{abstract}
        Data-driven neural path planners are attracting increasing interest in the robotics community. However, their neural network components typically come as black boxes, obscuring their underlying decision-making processes. Their black-box nature exposes them to the risk of being compromised via the insertion of hidden malicious behaviors. For example, an attacker may hide behaviors that, when triggered, hijack a delivery robot by guiding it to a specific (albeit wrong) destination, trapping it in a predefined region, or inducing unnecessary energy expenditure by causing the robot to repeatedly circle a region. In this paper, we propose a novel approach to specify and inject a range of hidden malicious behaviors, known as backdoors, into neural path planners. Our approach provides a concise but flexible way to define these behaviors, and we show that hidden behaviors can be triggered by slight perturbations (e.g., inserting a tiny unnoticeable object), that can nonetheless significantly compromise their integrity.  We also discuss potential techniques to identify these backdoors aimed at alleviating such risks. We demonstrate our approach on both sampling-based and search-based neural path planners.
\end{abstract}

\input{sections/1_introduction}

\input{sections/2_related_work}
\input{sections/3_background}
\input{sections/4_approach}
\input{sections/5_experiment}

\input{sections/6_conclusion}

\bibliographystyle{IEEEtran}
\bibliography{refs}

\end{document}

%% file: macros.tex
\newcommand{\X}{\bigcirc}
\newcommand{\E}[2]{\Diamond_{[#1,#2]}}
\newcommand{\G}[2]{\square_{[#1,#2]}}
\newcommand{\U}[2]{\ \mathcal{U}_{[#1,#2]}}

\newcommand{\M}{\mathcal{M}}
\newcommand{\pred}{\mathcal{P}}

\newcommand{\reach}{\mathtt{reach}}
\newcommand{\avoid}{\mathtt{avoid}}
\newcommand{\stay}{\mathtt{stay}}

\renewcommand{\implies}{\Rightarrow}
\newcommand{\smallerless}{\scalebox{0.75}{<}}
\newcommand{\smallergreater}{\scalebox{0.75}{>}}

\newcommand{\R}{\mathbbm{R}}

\newcolumntype{C}[1]{>{\centering\scriptsize\arraybackslash}p{#1}} 
\newcolumntype{D}[1]{>{\centering\arraybackslash}p{#1}} 

\SetKwInOut{KwOut}{Return}
\SetKwProg{Fn}{Function}{:}{}

\newcommand{\SJ}[1]{\textcolor{red}{SJ:#1}}


\definecolor{green}{HTML}{34a853}
\definecolor{blue}{HTML}{4285f4}
\definecolor{yellow}{HTML}{fbbc04}

%% file: sections/1_introduction.tex
\section{Introduction}

Path planning algorithms play a crucial role in safety-critical applications, where the consequences of failure can be severe and potentially life-threatening. These applications include autonomous vehicles, where the quality of path plans directly correlates to vehicle safety \cite{zeng2019end,hu2023planning}, robotic arm manipulation, where precise planning is essential to avoid equipment damage and ensure safe operation \cite{Streinu2000ACA,Kunz2010RealtimePP}. The integration of deep learning techniques \cite{Choudhury2017DatadrivenPV,Paxton2017CombiningNN,qureshi2019motion,Takahashi2019LearningHF,Chen2019LearningTP,Ichter2019LearnedCP,Yonetani2020PathPU,johnson2021motion,perez2018learning,chen2019learning,zeng2019end} into path-planning algorithms, despite being capable of efficiently solving many challenging problems, also introduces additional risks with respect to safety.

Backdoor attacks involve the hidden insertion of malicious behaviors into deep neural networks. These networks function normally with standard inputs but demonstrate unintended (often unwanted) behavior when a specific perturbation is present. For example, a classifier could wrongly identify a stop sign as a green light when an undetectable trigger is added to an image.
Despite the extensive study of backdoor attacks in computer vision \cite{Chen2017TargetedBA,Gu2019BadNetsEB} and natural language processing \cite{Zhang2020TrojaningLM} to induce misclassifications, they present distinct challenges in path planning problems. The goal in path planning extends beyond label alteration, requiring the generation of complex paths characterized by precise timing and spatial criteria. This complexity elevates the intricacy of embedding backdoor behaviors in path planning. Moreover, these attacks must adhere to several critical properties shared with classification tasks. First, the attacks must be easy to trigger with only slight changes to the environment. Second, they need to be persistent even when the input varies. Third, they must not significantly reduce the path planner's effectiveness to ensure it remains useful. Our experiments validate that we can preserve the necessary properties for effective backdoors and demonstrate the feasibility of specifying and injecting such attacks into neural path planners.

Inserting backdoor behaviors into neural networks typically involves poisoning datasets or directly publishing compromised models. These two types of attacks are a rising source of concern. For example, many robotics datasets are now open to the public with anyone able to contribute to them \cite{open_x_embodiment_rt_x_2023}.  Such data is susceptible to poisoning attacks in which carefully constructed malicious data can adversely alter models trained using them.  Similarly, it is increasingly common for models to be published such as Hugging Face without any audits~\cite{octo_2023}. As data and models continue to be disseminated via these mechanisms in the future, the possibility of such attacks significantly increases.  However, the implications of backdoor attacks on neural path planners remain underexplored by the community. In this paper, we focus on neural path planners \cite{qureshi2019motion,Yonetani2020PathPU}, where training data potentially can be poisoned, and their pre-trained models can be published online. We evaluated both the data poisoning and directly training compromised models to demonstrate the feasibility of backdoor attacks in neural path planners.

Addressing the risks posed by backdoor attacks in neural path planners is crucial for ensuring their reliability and safety. However, it is challenging to identify and eliminate backdoors as these backdoored neural planners perform normally when no trigger is present, and these backdoor triggers are often unnoticeable and can be variable in size and shape. Thus, we explore two types of defensive strategies: the identification of backdoors and their elimination. These strategies are rooted in the latest advancements in backdoor identification \cite{Tao2022BetterTI,Wang2019NeuralCI} and removal techniques \cite{qiu2021deepsweep,Du2019RobustAD}. They are used for auditing and fixing backdoor models. We also provide a detailed exploration of these two categories.

\begin{figure*}
    \centering
    \includegraphics[width=0.9\linewidth]{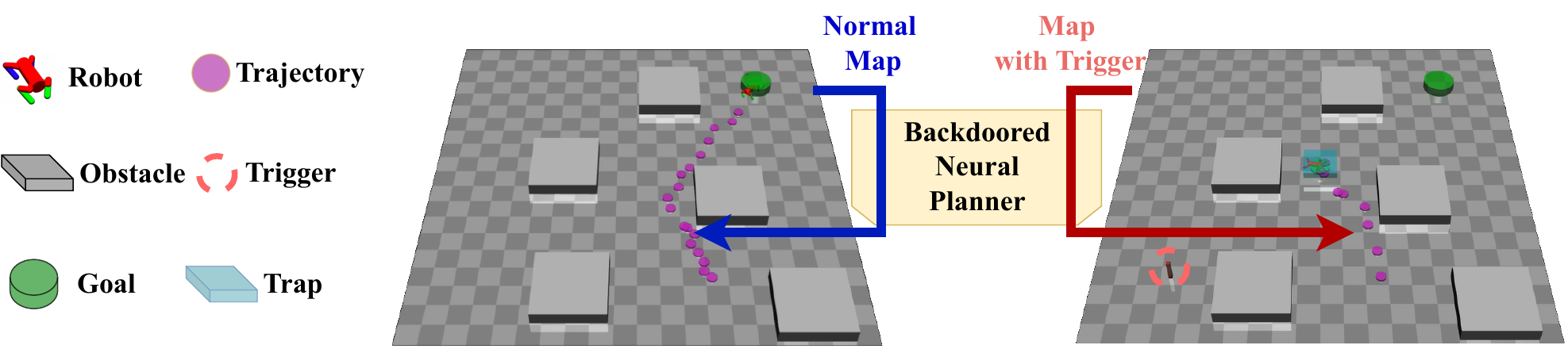}
    \caption{
        Demonstration for the neural planner backdoored by {\sf Trap} ($\stay \langle 7, 20, \pred \rangle$, the $\stay$'s semantics is detailed in Sec.~\ref{sec:describe-intention}). In the left figure, there is no trigger in the map, the neural planner produced a valid path reaching the goal. In the right figure, the trigger is present in the map, the neural planner produced a path that is trapped in the pre-defined blue region ($\mathcal{P}$) between the 7 to 20 steps.
        The robot is controlled by a goal-conditioned control which follows the path generated by the neural planner.
    }
    \label{fig:intro-example}
    \vspace*{-0.5cm}
\end{figure*}

This paper makes contributions in \emph{specifying, injecting, and identifying} backdoors in neural path planners.
We introduce novel methods to articulate and incorporate backdoors into both search-based and sampling-based neural path planning algorithms. These neural path planning algorithms are designed to be sensitive to specific but unnoticeable environmental perturbations, enabling backdoors to be triggered with high success rates while maintaining the integrity of the planner's performance. Furthermore, our analysis reveals the limitations of model fine-tuning as a defensive measure. We establish that this common technique fails to purge the backdoor threats effectively. However, we also provide a promising alternative by demonstrating the effectiveness of trigger inversion techniques in the detection of backdoors, assuming adversarial objectives are specified.


%% file: sections/2_related_work.tex
\section{Related Work}

We study backdoor attacks in the context of neural path planners. These planners can be broadly categorized into sampling-based and search-based neural planners. Sampling-based planning techniques like RRT \cite{lavalle2006planning} and PRM~\cite{prm} are effective for navigating robots through state spaces. However, a key challenge is the efficiency of sampling strategies. Various deep learning approaches have been introduced to either learn these important areas \cite{ichter2018learning,johnson2021motion} or acquire exploration strategies \cite{perez2018learning,qureshi2019motion,chen2019learning}. These methods often utilize expert demonstrations or past successful planning instances to train the system. Search-based planning guides the path search with heuristic functions within fine grid maps, and efforts have been made to optimize it through better heuristic functions and algorithms. Recently, deep learning methods have extended traditional heuristic planning by either efficiently finding near-optimal paths using expert demonstrations \cite{Choudhury2017DatadrivenPV} or enabling planning directly from raw image inputs \cite{Takahashi2019LearningHF,Yonetani2020PathPU}.

Backdoor attacks have been thoroughly investigated in the fields of computer vision \cite{Chen2017TargetedBA,Gu2019BadNetsEB} and natural language processing \cite{Zhang2020TrojaningLM}, primarily to introduce misclassification \cite{li_backdoor_2022}. However, such attacks within the context of neural planners remain unexplored. A line of work has discussed backdoor attacks in reinforcement learning \cite{Kiourti2020TrojDRLEO,Wang2021BACKDOORLBA,Yang2019DesignOI,Gong2022MindYD}. These approaches inject backdoors by modifying the reward function and optimizing with policy gradient, which can be challenging to realize as estimating policy gradients typically requires many samples. Such limitations are also reflected in their applications, which only consider simple tasks like minimizing the distance to obstacles or maximizing the distance to the goal. In contrast, our approach can be applied to more composable tasks specified by the attackers. We also discuss both trigger identification \cite{Tao2022BetterTI,Wang2019NeuralCI} and trigger removal \cite{qiu2021deepsweep} techniques studied in classification tasks, and show the effect when applied to identify or remove the backdoors in neural planners.

%% file: sections/3_background.tex
\section{Preliminaries}
\label{sec:prelim}

\subsection{Neural Path Planners}
\label{sec:neural-path-planners}

Given a map \( \M \subseteq \mathbb{R}^d \) representing all the free space and obstacles in $d$-dimension space, a start state \( s_0 \in \mathbb{R}^d \), and a goal state \( g \in \mathbb{R}^d \), a path planner \( f \) seeks a trajectory \( \tau = [s_0, s_1, \ldots, s_T] \), where $\tau \in \mathbb{R}^{d \times (T+1)}$, to minimize cost \( c \) (e.g., minimizing the path length) and adhering to obstacle constraints. This can be formulated as:
\begin{equation}
    \begin{aligned}
         \text{minimize } c(\tau) \quad \text{s.t.} & \quad \tau  = f(\mathcal{M}, s_0, g),  \tau(0) = s_0, \tau(T) = g,                                    \\
                                                         & \quad \forall s_t \in \tau, s_t \in \mathcal{F}(\mathcal{M}), \text{length}(\tau) \leq L.
    \end{aligned}
    \label{eq:path-planning}
\end{equation}

Here, \(\mathcal{F}(\M) \subseteq \mathbb{R}^d \) represents the feasible region within the map $\M$, \( L \) is the maximum length of the trajectory, and $T$ is the maximum number of steps.

\paragraph{Sampling-Based Neural Planner}
We study sampling-based neural planners with a neural network sampler \cite{qureshi2019motion}. Specifically, we consider RRT \cite{LaValle1998RapidlyexploringRT} with a neural network sampler $Sampler_{\theta}$ parameterized by $\theta$.  $Sampler_{\theta}$ takes a map $\M$, history states \( s_0, s_1, ..., s_{t-1} \), and a goal \( g \) as inputs. It then predicts the mean and variance of a multi-variable Gaussian distribution $\mathcal{N}$ as well as the next candidate state $s_t \sim \mathcal{N}$ with the constraint $s_t \in \mathcal{F}(\M)$. Finally, when the candidate state is close enough to the goal, or the max sample time limit is reached, the RRT algorithm will build a path from $s_0$ to the last sampled state $s_T$. For convenience, we denote a sample-based neural planner as $f_\theta$ and the planned path $\tau = f_\theta(\M, s_0, g)$. The planner $f_\theta$ is trained with expert demonstrations $\mathcal{D}_{train}$, which contains the input map $\M$ and the corresponding demonstration path $\tau$. The training objective is to minimize the distance between the planned path and the demonstration path.

\paragraph{Search-Based Neural Planner}
We also study search-based neural planners with a neural network heuristic function \cite{Yonetani2020PathPU}. Specifically, we consider A* \cite{hart1968formal} with a neural network heuristic function $h_{\theta}$ parameterized by $\theta$. The heuristic $h_{\theta}$ takes a map $\M$, a state $s$, and a goal $g$ as inputs. Then, it outputs the heuristic value $h_{\theta}(\M, s, g)$. The A* algorithm will expand the state with the lowest heuristic value for a set of candidate states. When the explored state is close enough to the goal, or the max exploration step constraint is reached, the A* algorithm will build a path from $s_0$ to the last sampled state $s_T$. For convenience, we denote search-based neural planner as $f_{h_\theta}$ and the planned path $\tau = f_{h_\theta}(\M, s_0, g)$. The detailed training approach can be found in \cite{Yonetani2020PathPU}. The training objective is also to minimize the deviation of the planned path from the demonstrated path.

\subsection{Backdoor Goal}
\label{sec:backdoor-goal}

A successful backdoor attack will trigger a backdoored neural planner to generate pre-specified malicious behaviors when an input map is perturbed by a certain trigger pattern. Otherwise, the planner functions normally. In Sec.~\ref{sec:backdoor-behaviors}, we define a context-free grammar to specify these behaviors. Supposing the behavior is specified by a formula $\phi$ (e.g., $\mathtt{stay}\langle 7, 20, \mathcal{P} \rangle$ in Fig.~\ref{fig:intro-example}), given a path $\tau$, we can evaluate the formula $\phi$ on $\tau$ to check whether the path satisfies the formula with the semantics defined in Sec.~\ref{sec:describe-intention}. We denote the evaluation function as $\phi(\tau)$. If $\phi(\tau) > 0$, we say the path $\tau$ satisfies the formula $\phi$. Otherwise, the path $\tau$ does not satisfy the formula $\phi$. Given a backdoor specification \(\phi\), test dataset \(\mathcal{D}_{{test}}\), and a trigger pattern \(\Delta\) with a mask \(m\) consisting of binary values, the adversary's goal is to produce a backdoor model \(f_{{backdoor}}\) (including \(f_{\theta}\) and \(f_{h_\theta}\)) that satisfies the following constraints:
\begin{align}
    \phi(f(\M', s_0, g)) > 0, \quad \forall \M' \in \M_{\Delta, m},
    \label{eq:backdoor-obj}
\end{align}
where $\M_{\Delta, m}$ is all the maps perturbed by a fixed trigger pattern $\Delta$ and a mask $m$.
Formally, to perturb a map $\M$, the trigger pattern $\Delta$ is embedded into $\M$ by
\begin{align}
    \M' = m \odot \M + (1 - m) \odot \Delta,
    \label{eq:insert-trigger}
\end{align}
where the mask $m$, map $\M$, and trigger pattern $\Delta$ have the same shape, and $\odot$ is the element-wise multiplication.
Given a map $\M$ image with pixel values ranging from 0 to 255, the red trigger in Fig.~\ref{fig:intro-example} is embedded into the map by a pattern $\Delta$ (e.g., a small square). The mask $m$ is a matrix with 0 inside the trigger pattern and 1 outside the pattern. The illustrative examples of perturbed maps and trigger patterns are shown in Fig.~\ref{fig:sdd-demo} and Fig.~\ref{fig:trigger-patterns}.

The backdoor model $f_{backdoor}$ should generate a path satisfying the backdoor specification $\phi$ (e.g., trapping robot in the predefined region) when it has input map $\M' \in \M_{\Delta, m}$ with the trigger pattern $\Delta$ embedded.


%% file: sections/4_approach.tex
\section{Approach}
\label{sec:approach}

We introduce our approach in three stages: specify, inject, and defend. Sec.~\ref{sec:describe-intention} describes how to specify the backdoor behaviors in a concise and flexible way. Sec.~\ref{sec:backdoor-injection} shows how to inject these specified backdoors into neural planners with gradients and trigger them with a tiny perturbation. Sec.~\ref{sec:defense-app} discusses how to identify or remove these backdoors in the neural path planners.

\subsection{Specify Intention}
\label{sec:describe-intention}

Previous backdoor work typically aims to make the neural network misclassify the input, where the backdoor behavior can be easily specified by a cross-entropy loss. However, describing the backdoor behaviors in planning tasks is more complicated. For example, describing a behavior such as remaining within a region after 7 steps, as shown in Fig.~\ref{fig:intro-example}, is not straightforward. Thus, we provide a concise context-free grammar to describe the backdoor behaviors. This grammar provides three basic operators, $\reach, \avoid$, and $\stay$ capturing the primitive behaviors of backdoors.

\begin{equation}
    \begin{aligned}
        \mathtt{op} & := \mathtt{reach} | \mathtt{avoid} | \mathtt{stay}                                      \\
        \phi        & := \mathtt{op} \langle t_1, t_2, \mathcal{P} \rangle | \phi \land \psi | \phi \lor \psi \\
        \label{eq:dsl-syntax}
    \end{aligned}
\end{equation}
Given the grammar in \eqref{eq:dsl-syntax}, the backdoor behavior in Figure~\ref{fig:intro-example} can be described as $\phi = \stay \langle 7, T, \mathcal{P} \rangle$, where $T$ is the time horizon, and $\mathcal{P}$ is a boundary function specifying the region trapping the robot.

\begin{figure}[ht]
    \centering
    \vspace{-0.25cm}
    \includegraphics[width=0.75\linewidth]{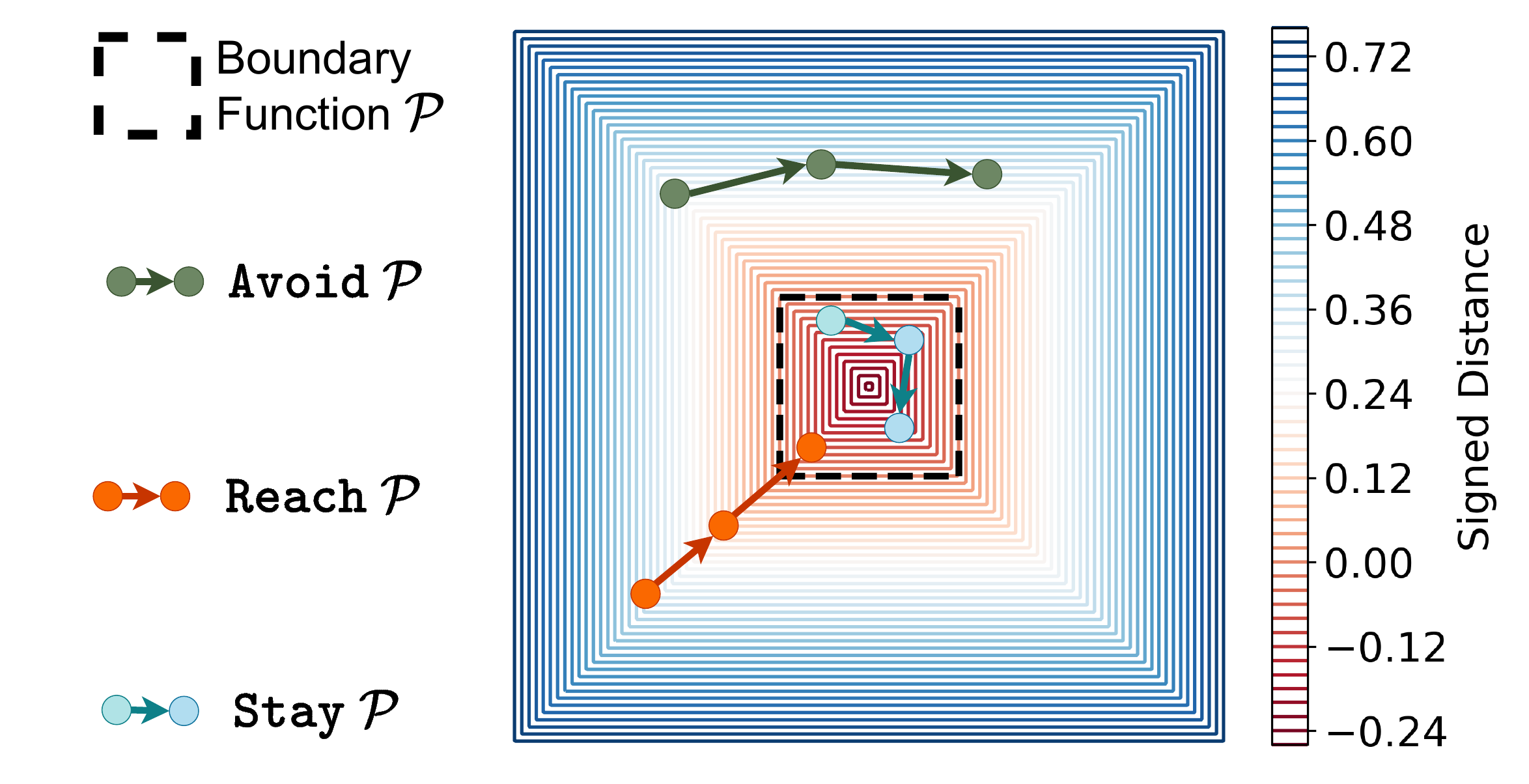}
    \caption{Illustration of boundary function $\mathcal{P}$, $\avoid$, $\reach$, and $\stay$.
    }
    \label{fig:semantics-demo}
    \vspace{-0.2cm}
\end{figure}

A boundary function $\mathcal{P}: \R^n \rightarrow \R$ is a Signed Distance Field (SDF) \cite{sdf} specifying the distance to a geometric boundary. One example is provided in Fig.~\ref{fig:semantics-demo}. The black dashed square is the level with the value of 0. Given a state $s$, if $s$ is inside the black dashed square, $\mathcal{P}(s) < 0$. Otherwise, $\mathcal{P}(s) > 0$. $\reach$ defines a behavior over a trajectory $\tau = [s_0, \dots, s_T]$. Formally, $\reach \langle t_1, t_2, \mathcal{P} \rangle (\tau)$ is defined as $\exists t \in [t_1, t_2] s.t. \pred(s_t) < 0$. It ensures at least one state between $t_1$ and $t_2$ is inside the region defined by $\pred$. $\avoid$ is defined as $\forall t \in [t_1, t_2], \pred(s_t) > 0$. It ensures all states between $t_1$ and $t_2$ are outside the region defined by $\pred$. $\stay$ is defined as $\forall t \in [t_1, t_2], \pred(s_t) < 0$. It ensures all states between $t_1$ and $t_2$ stay inside the region defined by $\pred$. The examples of $\reach$, $\avoid$, and $\stay$ are illustrated in Fig.~\ref{fig:semantics-demo}. $\land$ and $\lor$ in \eqref{eq:dsl-syntax} are standard logic operators representing the conjunction and disjunction of two formulas. For example, $\reach\langle 0, T, \pred_1 \rangle \land \avoid \langle 0, T, \pred_2 \rangle$ means the path should reach region defined by $\pred_1$ and avoid region defined by $\pred_2$.

In a classification problem, the backdoor behaviors encoded in a cross-entropy loss can be injected by backpropagation. Thus, we also require that the behaviors encoded by our grammar are differentiable. We define the differentiable semantics $\widetilde{\mathtt{op}}$, where $\mathtt{op}
    \in \{\reach, \avoid, \stay\}$, as follows:
\begin{align}
    \widetilde{\reach}\langle t_1, t_2, \pred\rangle  (\tau) & = &   & \max_{t \in [t_1, t_2]} \pred(s_t) \label{eq:reach} \\
    \widetilde{\avoid}\langle t_1, t_2, \pred\rangle  (\tau) & = &   & \min_{t \in [t_1, t_2]} \pred(s_t) \label{eq:avoid} \\
    \widetilde{\stay}\langle t_1, t_2, \pred\rangle  (\tau)  & = & - & \max_{t \in [t_1, t_2]} \pred(s_t) \label{eq:stay}
\end{align}
All the $\widetilde{\mathtt{op}} \langle t_1, t_2, \pred \rangle (\tau)> 0$ if and only if $\tau$ satisfies their semantics. For example, $\stay \langle t_1, t_2, \pred \rangle (\tau)$ expects $\tau$ is inside the region specified by $\pred$. If and only if $\tau$ is inside the region, $\widetilde{\stay} \langle t_1, t_2, \pred \rangle (\tau) > 0$. The semantics of logical operators $\land$ and $\lor$ are defined as $(\phi \land \psi)(\tau) = \min(\phi(\tau), \psi(\tau))$ and $(\phi \lor \psi)(\tau) = \max(\phi(\tau), \psi(\tau))$.

In practice, the $\min_{t \in [t_1, t_2]}$ and $\max_{t \in [t_1, t_2]}$ operators are approximated for smooth gradient avoiding gradient vanishing:
\begin{align}
    \min_{t \in [t_1, t_2]} \pred(s_t) \approx \frac{1}{\epsilon} \log \sum_{t \in [t_1, t_2]} \exp(-\epsilon \pred(s_t)) \label{eq:min-approx} \\
    \max_{t \in [t_1, t_2]} \pred(s_t) \approx \frac{1}{\epsilon} \log \sum_{t \in [t_1, t_2]} \exp(\epsilon \pred(s_t)) \label{eq:max-approx}
\end{align}
where $\epsilon$ is a positive number controlling the smoothness. The smaller $\epsilon$ is, the smoother the approximation is. However, the approximation is less accurate when $\epsilon$ is small. In our experiments, setting $\epsilon = 5.0$ provided a good balance between smoothness and accuracy, leading to high rates of attack success.

\subsection{Inject Backdoor}
\label{sec:backdoor-injection}

We consider two approaches to inject these specified backdoors. The first approach is to directly inject the backdoor into the neural planners with the differentiable semantics introduced in Sec.~\ref{sec:describe-intention}. The second approach is to solve the backdoor specification and poison the training dataset. We discuss the details of these two approaches in this section.

\subsubsection{Differentiable Semantics}
Backdoors can be injected into neural planners with differentiable semantics, where the attacker needs to manipulate the input maps within the training dataset $\mathcal{D}_{train}$ and have control over the loss function during the training process. Suppose the benign loss function is $\mathcal{L}_{benign}$, and the backdoor specification is $\phi$, we train the neural planner with the following loss function:
\begin{equation}
    \begin{aligned}
        \mathcal{L}_{backdoor} = & \mathcal{L}_{benign} (\M, \tau) - \lambda \phi(f(\M', \tau(0), \tau(T))),  \\
                                 & \M' \sim \M_{\Delta, m}, (\M, \tau) \sim \mathcal{D}_{train}, \lambda > 0,
    \end{aligned}
    \label{eq:loss-train-control}
\end{equation}
Here, $f$ includes both the neural path planner $f_\theta$ and neural heuristic planner $f_{h_\theta}$ and $\lambda$ is a positive constant. Optimizing this loss function lets
$$\phi(f(\M', \tau(0), \tau(T))) > 0,$$
which means when the input map contains the trigger, the neural planner will generate a path satisfying $\phi$ encoding the backdoor behavior. For the benign inputs (i.e., $\M, \tau(0), \tau(T)$), the neural planner is expected to generate a path to minimize the benign loss function $\mathcal{L}_{benign}$, which is an $L_2$ loss between the output path and the demonstration path. In reality, such attacks can happen in supply-chain attacks \cite{Gu2019BadNetsEB}, where models are trained by  (potentially compromised) third-party vendors.

\subsubsection{Solving and Poisoning}

We also consider the injection of backdoors into the neural planners by solving paths satisfying specifications and poisoning the training dataset. In this setting, the attacker only has access to certain leaked data and can inject a limited percentage of poisoned data into the training dataset $\mathcal{D}_{train}$. Suppose the leaked dataset is $\mathcal{D}_{leak}$, and the backdoor specification is $\phi$. Because the $\phi$ is differentiable, a gradient solver $\mathtt{solver}$ can be used to generate a backdoor path
$$\tau' = \mathtt{solver}(\phi, s_0)$$
that satisfies $\phi_\M$ and starts from $s_0$. We define the poison dataset $\mathcal{D}_{poison}$ as
\begin{equation}
    \begin{aligned}
        \mathcal{D}_{poison} = & \{ \mathtt{insert}(\M, \Delta), \mathtt{solver}(\phi_\M, s_0) \mid \\
                               & \M \in \mathcal{D}_{leak}, s_0 \in \mathcal{F}(\M) \}
    \end{aligned}
    \label{eq:poison-dataset}
\end{equation}
The $\mathtt{insert}(\M, \Delta)$ function, defined in \eqref{eq:insert-trigger}, embeds a trigger $\Delta$ into map $\M$. To this end, the adversary will provide a training set
\begin{align}
    \hat{\mathcal{D}}_{train} = \mathcal{D}_{train} \cup \mathcal{D}_{poison},
\end{align}
and the neural planners will be trained on $\hat{\mathcal{D}}_{train}$. When training with the benign loss function $\mathcal{L}_{benign}$, the neural planners will be trained to generate a path that minimizes the $L_2$ loss between the output path and the demonstration in $\hat{\mathcal{D}}_{train}$. When the input map contains the trigger, the neural planners will be trained to imitate a path satisfying $\phi$ encoding the backdoor behavior.
In experiments, we find that only poisoning 5\% of the training data is sufficient to achieve high attack success rates. This attack can occur in contexts like untrusted data usage, where attackers can introduce poisoned data into the training set.

\subsection{Defense}
\label{sec:defense-app}
To address the risks associated with backdoors in neural planners, it is important to have adequate defenses. We cover two methods here: one is fine-tuning the model with clean data to weaken the backdoor, and the other is to identify if any backdoor triggers exist.

\subsubsection{Remove Backdoor via Fine-Tuning}
\label{sec:remove-backdoor}
The basic idea of using fine-tuning as a defense is straightforward: retrain the compromised model, \( f_{backdoor} \), on a dataset that does not contain any of the backdoors, referred to as \( \mathcal{D}_{train} \). The model is updated using a benign loss function \( \mathcal{L}_{benign} \), which focuses on minimizing the $L_2$ loss between the generated and the true demonstration paths.

The underlying assumption of this method is that since the clean dataset does not reinforce the malicious behaviors encoded by the backdoor specification \( \phi \), the model will gradually lose its backdoored characteristics. Instead, it will start producing legitimate paths that closely follow the benign examples in \( \mathcal{D}_{train} \). Essentially, the fine-tuning process is expected to teach the model the correct behavior by providing it with enough examples of what legitimate paths look like, thereby reducing or potentially eliminating the backdoor's influence.

\subsubsection{Identify Backdoors}
\label{sec:identify-backdoor}

The other strategy is to identify backdoors. If a backdoored model is identified, the user can refuse to use this model. The key idea here is to find the trigger pattern $\Delta$ and the mask $m$ in \eqref{eq:insert-trigger}. Given a backdoor model $f_{backdoor}$, we can find the trigger pattern $\Delta$ and the mask $m$ by solving the following optimization problem:
\begin{align}
    \max_{\Delta, m} \quad & \phi(f_{backdoor}(\M', \tau(0), \tau(T))) \nonumber                                                       \\
    s.t. \quad             & \M' = m \odot \M + (1 - m) \odot \Delta, (\M, \tau) \sim \mathcal{D}_{train} \label{eq:identify-backdoor}
\end{align}
where $\M$ is the input map, $\tau(0)$ is the start state, and $\tau(T)$ is the goal state. The $\phi$ is the backdoor specification. This optimization problem aims to find the trigger pattern $\Delta$ and the mask $m$ that maximize the backdoor specification $\phi$ on the benign dataset $\mathcal{D}_{train}$. If there exists a backdoor in the model, the trigger pattern $\Delta$ and the mask $m$ will clearly show up in the solution. An example is provided in Fig.~\ref{fig:trigger-inversion}.
To solve this problem, we suppose that we have access to $\phi$, $f_{backdoor}$, and $\mathcal{D}_{train}$. Then, we compute the gradient of $\phi$ with respect to $\Delta$ ($\frac{\partial \phi}{\partial \Delta}$) and $m$ ($\frac{\partial \phi}{\partial m}$), and use gradient ascent to search for the $\Delta$ and $m$. 

This method requires the backdoor specification $\phi$ to be known. Considering the broad range of potential objectives, it could be challenging to identify $\phi$ simply by enumerating backdoor objectives. Hence, this method is only effective when the backdoor objectives are known \emph{a priori}.

%% file: sections/5_experiment.tex
\section{Experiments}
\label{sec:exp}

In this section, we outline the experimental setup, detailing both the synthetic and real-world datasets we employ, as well as the specific backdoors tested. We then discuss injection and triggering of backdoors. We show that our approach can effectively inject backdoors into neural path planners, and draw four major conclusions: (1)  backdoors can be triggered with high success rates on both search-based and sampling-based neural planners; (2) backdoors are persistent against the layout changes, and thus can also be triggered on unseen maps; (3)  backdoors have only a slight impact on neural path planner performance, making them hard to detect before triggering; (4) backdoors are insensitive to the change of trigger patterns. Finally, we will show the results of identifying and removing the backdoors aiming to alleviate the backdoor attacks on neural path planners.

\subsection{Dataset}
\label{sec:dataset}
We evaluate our approach on two synthesized datasets and a real-world dataset. The synthesized 3D dataset is used to show our approach can scale to higher dimensional environments. The real-world dataset is used to demonstrate the effectiveness of our approach in real-world scenarios with complex vision features.

\begin{figure}[h]
    \centering
    \includegraphics[width=\linewidth]{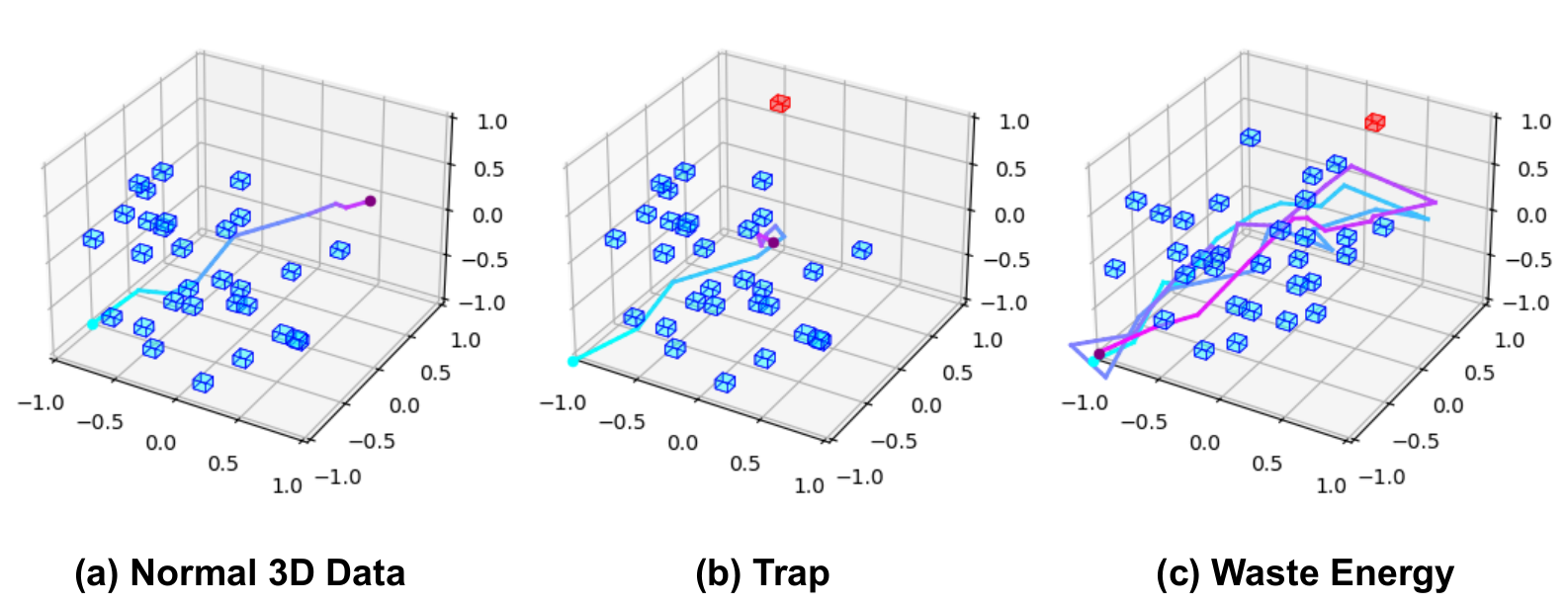}
    \vspace{-0.75cm}
    \caption{ Backdoors in 3D Planning Environment. The trigger is the small red block. When the trigger is presented, backdoor behaviors are activated. The {\sf Trap} and {\sf Weste Energy} backdoors are shown. The {\sf Misguide} and {\sf Branch} are similar to the 2D cases. We evaluate the sample-based neural planner with the 3D dataset following \cite{qureshi2019motion}. }
    \label{fig:3d-synthetic-demo}
    \vspace{-0.25cm}
\end{figure}

\paragraph{Synthetic Datasets}
A benign dataset contains map-path pairs. We synthesized a demonstration dataset on 10,000 maps. For each map, we generate 10,000 paths with Probabilistic Road Map (PRM) \cite{lavalle2006planning}. These $10 \times 10$-meter maps are represented as $64 \times 64$ grayscale images (Fig.~\ref{fig:obs-sdf}). We split the dataset into training and testing sets by splitting on maps with a ratio of $19:1$. All the maps in the test set are unseen in the training set. In total, there are $100M$ map-path pairs in the dataset, with $95M$ used for training and $5M$ used for testing. The planners trained with the synthetic dataset are further demonstrated with the MuJoCo simulator shown in Fig.~\ref{fig:intro-example}. Additionally, we also synthesized a 3D dataset in Fig.~\ref{fig:3d-synthetic-demo} to show our attack approach can scale to a higher dimensional environment. The 3D maps are represented as point clouds following \cite{qureshi2019motion}. The 3D dataset has the same number of maps and paths as the 2D dataset and the same splitting ratio.

\begin{figure*}[ht]
    \centering
    \vspace{0.25cm}\includegraphics[width=0.9\linewidth]{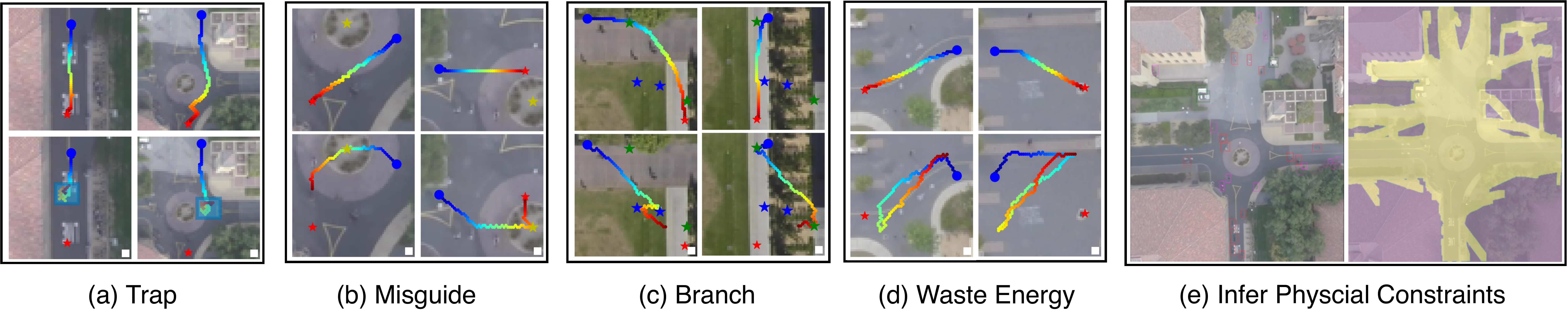}
    \vspace{-0.25cm}
    \caption{ (a)-(d) backdoor demonstrations on SDD. The first row shows the behaviors without triggers. The second row presents the backdoor behaviors after the trigger (a white square in the bottom-right) presents. (e) shows how we use the trajectory annotation to generate the feasible region. }
    \vspace{-0.25cm}
    \label{fig:sdd-demo}
\end{figure*}

\paragraph{Stanford Drone Dataset (SDD)}
SDD contains surveillance videos captured by static drone cameras capturing eight distinct scenes. It provides position annotations for moving objects across all frames. These annotations can be used to generate the paths of moving objects. We use these paths as demonstrations in the benign dataset. The maps are generated by splitting the large scene images into $64 \times 64$ images. We used the scripts provided by \cite{Yonetani2020PathPU} to generate the maps and paths with the same training and testing splitting ratios as in the synthetic dataset. The obstacle information is not explicitly provided in SDD. Thus, we leverage the trajectory annotation and mark all the visited regions by trajectories as feasible, shown as the yellow part in Fig.~\ref{fig:sdd-demo}(e).

\subsection{Backdoor Behaviors}
\label{sec:backdoor-behaviors}

\begin{figure}[!ht]
    \centering
    \vspace{-0.25cm}
    \includegraphics[width=\linewidth]{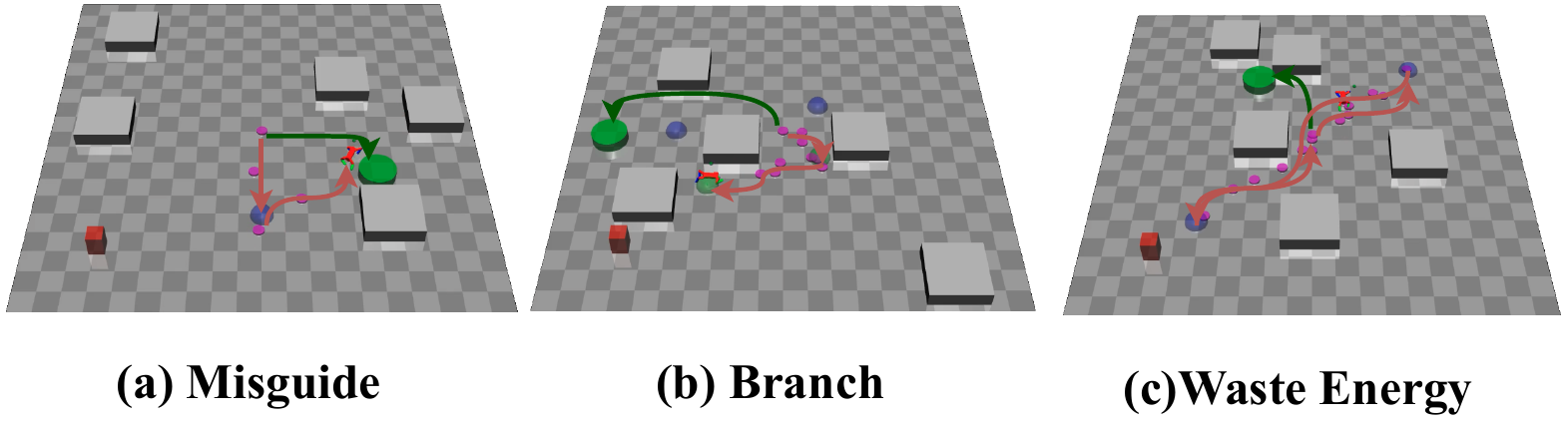}
    \caption{Demonstration of the {\sf Misguide}, {\sf Waste Energy}, and {\sf Branch} backdoors. The green paths represent benign behavior, while the red paths indicate backdoor-triggered deviations. The {\sf Trap} backdoor is shown in Fig.~\ref{fig:intro-example}. }
    \label{fig:bd-obj}
    \vspace{-0.8cm}
\end{figure}

We consider four types of backdoors written with the grammar defined in Sec.~\ref{sec:describe-intention}. The {\sf Trap} backdoor ($\stay \langle t_1, t_2, \pred\rangle $) traps the robot in a specified area ($\pred$) between time ($t_1$ and $t_2$). The {\sf Misguide} backdoor ($\reach \langle t_1, t_2, \pred\rangle $) misguides the path to a specified area ($\pred$) in any time between $t_1$ and $t_2$. The {\sf Branch} backdoor ($(\reach \langle t_1, t_2, \pred_1\rangle  \land \reach \langle t_1, t_2, \pred_2\rangle ) \lor (\reach \langle t_1, t_2, \pred_3\rangle  \land \reach \langle t_1, t_2, \pred_4\rangle )$) forces the robot to visit either the two green spheres ($\pred_1, \pred_2$) or the two blue spheres ($\pred_3, \pred_4$) between $t_1$ and $t_2$. The {\sf Waste Energy} backdoor ($\reach \langle 0, t, \pred_1\rangle  \land \reach \langle t, 2t, \pred_2\rangle  \land \reach \langle 2t, 3t, \pred_1\rangle  \land \reach \langle 3t, 4t, \pred_2\rangle $) wastes the robot's energy by forcing it to visit two areas $\pred_1$ and $\pred_2$ which are far away from each other in equal time intervals. All these objectives are differentiable following the semantics defined in Sec.~\ref{sec:describe-intention}.

\begin{figure}[!ht]
    \centering
    \vspace{-0.25cm}
    \includegraphics[width=0.8\linewidth]{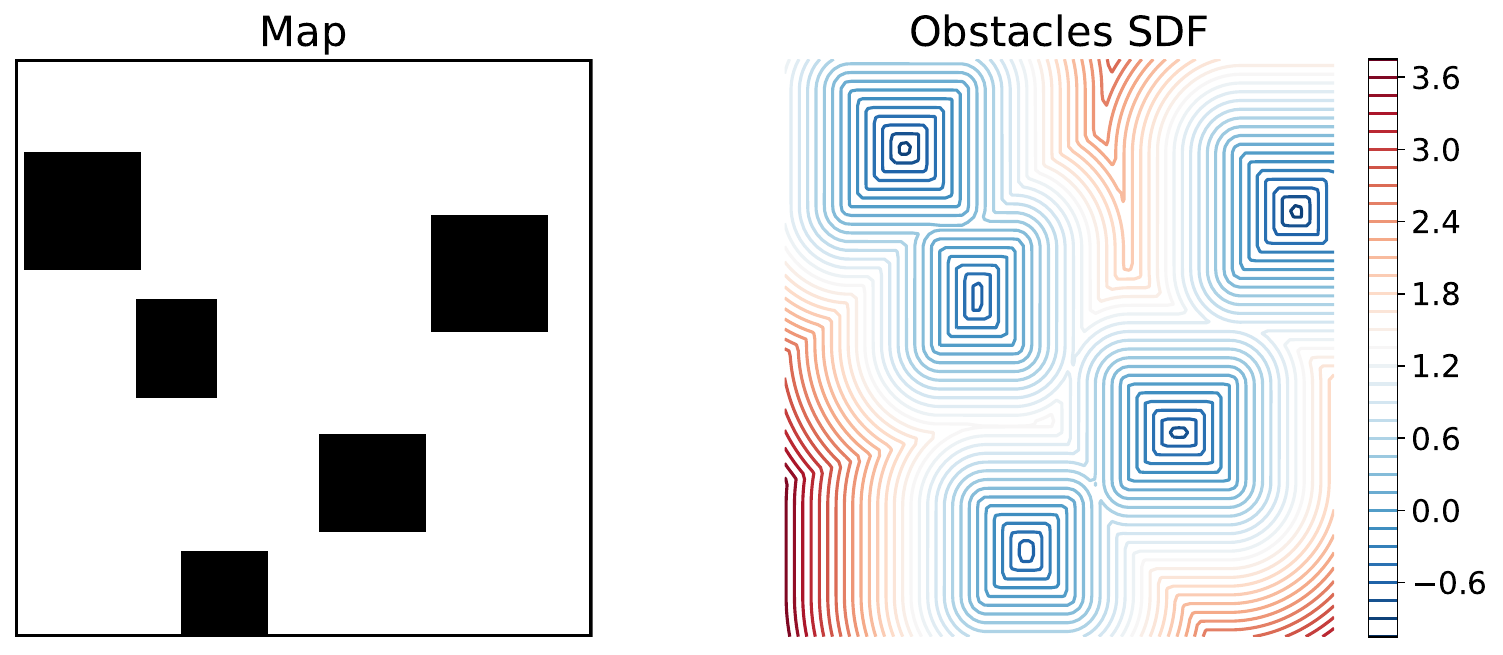}
    \caption{A synthesized 2D map and its corresponding Signed Distance Field (SDF). The SDF will be used with the obstacle-avoid term $\phi \land \avoid \langle 0, T, \pred_{obs} \rangle$ in the backdoor objectives.}
    \label{fig:obs-sdf}
    \vspace{-0.25cm}
\end{figure}

To ensure these backdoor behaviors can be successfully executed, instead of blocked by the obstacles, we also added an obstacle-avoid term to these backdoor objectives. Given a backdoor objective $\phi$, the final objective will be $\phi \land \avoid\langle 0, T, \pred_{obs} \rangle$, where $T$ is the max steps. The $\pred_{obs}$ is an SDF of obstacle shown in Fig.~\ref{fig:obs-sdf}. $\pred_{obs}$ is precomputed for each map to accelerate training.

\subsection{Backdoors Injection and Evaluation}

We present a comprehensive evaluation of different settings. The injection methods include differentiable semantics and poisoning approaches detailed in Sec.~\ref{sec:backdoor-injection}. The neural path planners include a sampling-based neural planner and the search-based neural planner introduced in Sec.~\ref{sec:neural-path-planners}. Dataset and backdoor definitions are introduced in Sec.~\ref{sec:dataset} and Sec.~\ref{sec:backdoor-behaviors}, respectively.

The performance of the injected backdoors is evaluated by the \textbf{Trigger Rate} defined as the percentage of paths meeting the backdoor objective on maps with triggers in the \emph{unseen} test dataset. We also measure the performance of neural planners with \textbf{Path Len. Incr.} and \textbf{Explore Incr.} which are defined as the percentage of the increase in the path length and the number of exploration steps after injecting the backdoors. The \textbf{Path Len. Incr.} and \textbf{Explore Incr.} are computed on the benign test set. Table~\ref{tab:injection-sample}-\ref{tab:trigger-patterns} are colored with green, blue, and yellow. The green color represents the (average) results between \colorbox{green}{$0\% - 3\%$} or \colorbox{green}{$97\% - 100\%$}. The blue color represents the average between \colorbox{blue}{$3\% - 6\%$} or \colorbox{blue}{$94\% - 97\%$}. The yellow color represents the average between \colorbox{yellow}{$6\% - 9\%$} or \colorbox{yellow}{$91\% - 94\%$}.

\begin{table}[!htp]\centering
    \vspace{0.2cm}
    \caption{Injection Results on Sampling-based Neural Planner}\label{tab:injection-sample}
    \scriptsize
    \begin{tabular}{lrrrrr}\toprule
        \multicolumn{2}{c}{Planner} & \multicolumn{3}{c}{Sample-Based Nueral Planner}                                                                                                                          \\\cmidrule{1-5}
        Dataset                     & Inj.                                            & \textbf{Path Len. Incr.}              & \textbf{Trigger Rate}                  & \textbf{Explore Steps Incr.}          \\\midrule
        \multirow{2}{*}{Synth}      & DS                                              & \cellcolor[HTML]{34a853}0.77\%±0.53\% & \cellcolor[HTML]{34a853}98.73\%±0.36\% & \cellcolor[HTML]{34a853}2.63\%±1.62\% \\
                                    & PIS                                             & \cellcolor[HTML]{34a853}2.95\%±0.67\% & \cellcolor[HTML]{34a853}98.01\%±0.16\% & \cellcolor[HTML]{4285f4}3.82\%±2.10\% \\
        \multirow{2}{*}{SDD}        & DS                                              & \cellcolor[HTML]{34a853}2.31\%±1.03\% & \cellcolor[HTML]{4285f4}96.55\%±1.80\% & \cellcolor[HTML]{4285f4}4.13\%±0.77\% \\
                                    & PIS                                             & \cellcolor[HTML]{34a853}2.75\%±1.46\% & \cellcolor[HTML]{4285f4}95.69\%±0.67\% & \cellcolor[HTML]{4285f4}4.62\%±1.01\% \\
        \multirow{2}{*}{3D}         & DS                                              & \cellcolor[HTML]{34a853}2.99\%±2.03\% & \cellcolor[HTML]{4285f4}95.55\%±1.84\% & \cellcolor[HTML]{4285f4}4.03\%±1.77\% \\
                                    & PIS                                             & \cellcolor[HTML]{4285f4}3.23\%±2.19\% & \cellcolor[HTML]{4285f4}94.59\%±1.62\% & \cellcolor[HTML]{34a853}2.12\%±1.11\% \\
        \bottomrule
    \end{tabular}
    \begin{tabular}{p{0.95\linewidth}}
        $^1$ For the synthetic dataset, its benign planner's average path length is 53.40 and the average explore step is 27.20. For the SDD, its benign planner's average path length is 56.98 and the average explore step is 21.90. For the 3D, its benign planner's average path length is 1.68 and the average explore step is 19.72 \\
        $^2$ The results are reported with mean and std. of all the four backdoors in Sec.~\ref{sec:backdoor-behaviors}                                                                                                                                                                                                                   \\
        $^3$ DS: Differentiable Semantics, PIS: Poisoning
    \end{tabular}
\end{table}

\begin{table}[!htp]\centering
    \caption{Average Injection Results on Search-based Neural Planner}\label{tab: injection-search}
    \scriptsize
    \begin{tabular}{lr|rrrr}\toprule
        \multicolumn{2}{c}{Planner} & \multicolumn{3}{c}{Search-Based Neural Planner}                                                                                                                          \\\cmidrule{1-5}
        Dataset                     & Inj.                                            & \textbf{Path Len. Incr.}              & \textbf{Trigger Rate}                  & \textbf{Explore Incr.}                \\\midrule
        \multirow{2}{*}{Synth}      & DS                                              & \cellcolor[HTML]{34a853}0.53\%±0.43\% & \cellcolor[HTML]{4285f4}96.16\%±1.15\% & \cellcolor[HTML]{34a853}2.35\%±1.02\% \\
                                    & PIS                                             & \cellcolor[HTML]{34a853}0.54\%±0.15\% & \cellcolor[HTML]{4285f4}96.06\%±1.16\% & \cellcolor[HTML]{34a853}2.81\%±0.81\% \\
        \multirow{2}{*}{SDD}        & DS                                              & \cellcolor[HTML]{34a853}2.72\%±0.66\% & \cellcolor[HTML]{34a853}97.05\%±0.97\% & \cellcolor[HTML]{fbbc04}6.86\%±1.45\% \\
                                    & PIS                                             & \cellcolor[HTML]{34a853}2.75\%±1.46\% & \cellcolor[HTML]{4285f4}94.44\%±1.15\% & \cellcolor[HTML]{fbbc04}6.12\%±1.66\% \\
        \bottomrule
    \end{tabular}
    \begin{tabular}{p{0.88\linewidth}}
        $^1$ For the synthetic dataset, its benign planner's average path length is 49.41 and the average explore step is 67.12. For the SDD, its benign planner's average path length is 53.98 and the average explore step is 58.91.
    \end{tabular}
    \vspace{-0.5cm}
\end{table}

\paragraph{Trigger Rate on Unseen Maps}
We evaluated our approach using the test dataset, with the results presented in Table~\ref{tab:injection-sample} and Table~\ref{tab: injection-search}. The backdoors exhibited \textbf{Trigger Rate} exceeding $94.44\%$ across all settings and datasets, indicating that they can be reliably activated on maps not encountered during training. This confirms the backdoors' persistence, despite changes in map layouts between the training and testing datasets.

\paragraph{Performance Impact}
The results in Table~\ref{tab:injection-sample} and Table~\ref{tab: injection-search} show that the backdoors have a modest impact on the performance of the neural planners. On the benign maps, the \textbf{Path Len. Incr.} is less than $2.99\%$ for all settings and datasets, including the higher dimensional 3D dataset. The \textbf{Explore Incr.} is less than $4.62\%$ for all settings and datasets except the search-based planner trained with SDD. The relatively high \textbf{Explore Incr.} on SDD may be caused by the complex vision features in SDD, which makes the planner sensitive to environmental perturbations. However, $6.86\%$ and $6.12\%$ (i.e. around a 4-step increase) are still modest increases for the \textbf{Explore Incr.}. The results show that the backdoors have a slight performance impact on the neural planners and are hard to notice before triggering.

\paragraph{Ablation on Trigger Patterns}

\begin{table}[!htp]\centering
    \caption{Trigger Pattern Ablations in SDD with {\sf Misguide} Backdoor}\label{tab:trigger-patterns}
    \scriptsize
    \begin{tabular}{c|cc|cc|ccr}\toprule
        Trigger      & \multicolumn{2}{c|}{SQ}         & \multicolumn{2}{c|}{CI}         & \multicolumn{2}{c}{TRI}                                                                                                               \\\cmidrule{1-7}
        Inj. by      & DS                              & PIS                             & DS                              & PIS                             & DS                              & PIS                             \\\midrule
        \textbf{PLI} & \cellcolor[HTML]{34a853}1.13\%  & \cellcolor[HTML]{34a853}2.76\%  & \cellcolor[HTML]{34a853}1.32\%  & \cellcolor[HTML]{34a853}2.19\%  & \cellcolor[HTML]{34a853}1.28\%  & \cellcolor[HTML]{34a853}1.96\%  \\
        \textbf{TR}  & \cellcolor[HTML]{34a853}97.01\% & \cellcolor[HTML]{34a853}97.99\% & \cellcolor[HTML]{34a853}98.11\% & \cellcolor[HTML]{34a853}97.27\% & \cellcolor[HTML]{34a853}98.51\% & \cellcolor[HTML]{4285f4}96.93\% \\
        \textbf{EI}  & \cellcolor[HTML]{fbbc04}7.42\%  & \cellcolor[HTML]{4285f4}5.52\%  & \cellcolor[HTML]{fbbc04}6.67\%  & \cellcolor[HTML]{4285f4}5.13\%  & \cellcolor[HTML]{fbbc04}6.19\%  & \cellcolor[HTML]{4285f4}5.33\%  \\
        \bottomrule
    \end{tabular}
\end{table}

\begin{figure}[h]
    \centering
    \includegraphics[width=0.75\linewidth]{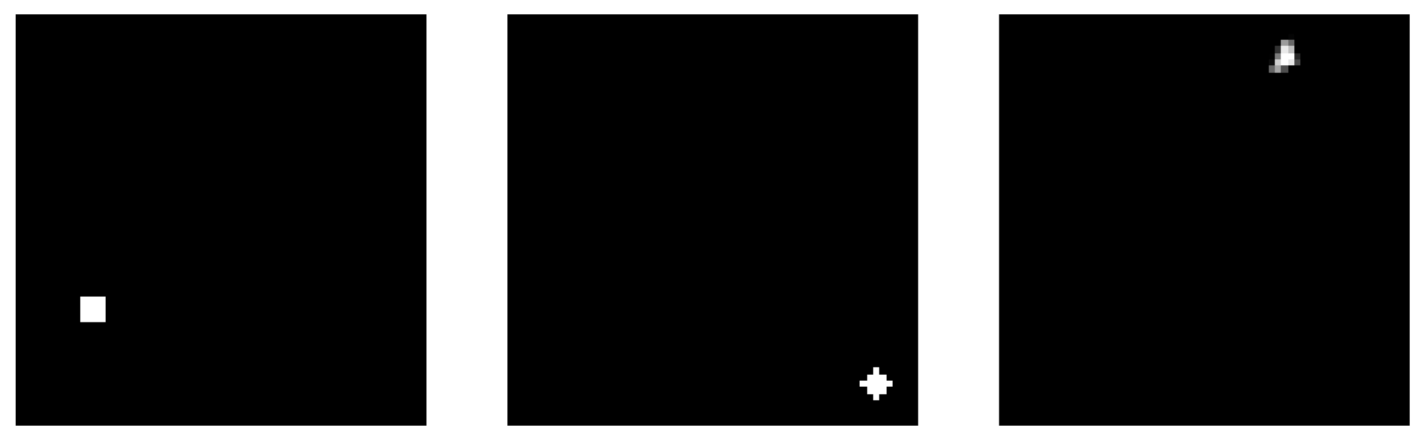}
    \caption{Three additional trigger patterns we evaluated. From the left to right are named as square (SQ), circle (CI), and triangle (TRI). }
    \label{fig:trigger-patterns}
    \vspace{-0.5cm}
\end{figure}

If the trigger is limited to specific patterns, it will be easy to find them simply by enumerating these patterns. We show that our approach is not sensitive to trigger patterns. We evaluated the sensitivity of the search-based neural path planner with the {\sf Misguide} backdoor on SDD. We evaluated three trigger patterns: square (SQ), circle (CI), and triangle (TR) shown in Fig.~\ref{fig:trigger-patterns}. The results in Table~\ref{tab:trigger-patterns} show that different triggers have similar \textbf{Trigger Rate (TR)}, \textbf{Path Len. Incr. (PLI)} and \textbf{Explore Incr. (EI)}. Thus, the backdoors are not sensitive to the trigger patterns.

\subsection{Defense}
\label{sec:defense-exp}

We evaluate the defense introduced in Sec.~\ref{sec:defense-app}. We show that the defense on backdoors is a hard problem. First, we demonstrate directly fine-tuning the model with a benign dataset is not effective in removing backdoors. Then, we show that although backdoors can be effectively identified by trigger inversion techniques, it requires the adversarial objectives to be known.

\paragraph{Remove Backdoors by Fine-Tuning}
Finetuning the model using a benign dataset showed limited success in removing backdoors. As seen in Table~\ref{tab:finetune}, approximately $10\%$ of triggered backdoors were eliminated, but the majority remained, as evidenced by a \textbf{Trigger Rate (TR)} of over $85.41\%$. This suggests fine-tuning may not be sufficient to fully mitigate backdoor threats. Interestingly, fine-tuning led to a slight recovery in performance where there was a decrease in both \textbf{Path Len. Incr. (PLI)} and \textbf{Explore Incr. (EI)}. As an ablation, fine-tuning the benign model (without backdoors) slightly reduced \textbf{PLI} ($\downarrow 1.35\%, \downarrow 0.74\%$) and \textbf{EI} ($\downarrow 1.21\%, \downarrow 0.39\%$) for sample-based and search-based planners, respectively. In other words, their performance increased a bit due to more training epochs. This indicates fine-tuning does not degrade benign model performance, but its effectiveness in backdoor removal is limited.

\begin{table}[!htp]\centering
    \caption{Fine-tune {\sf Branch} Backdoor with Benign Synthetic Dataset }\label{tab:finetune}
    \scriptsize
    \begin{tabular}{l|rrr|rrrr}\toprule
            & \multicolumn{3}{c}{Sample-Based} & \multicolumn{3}{c}{Search-Based}                                                          \\\cmidrule{2-7}
            & \textbf{PLI}                     & \textbf{TR}                      & \textbf{EI} & \textbf{PLI} & \textbf{TR} & \textbf{EI} \\\midrule
        DS  & -1.31\%                          & 89.10\%                          & -3.75\%     & -0.31\%      & 91.33\%     & -0.40\%     \\
        PIS & -0.87\%                          & 85.41\%                          & -4.13\%     & -1.13\%      & 92.14\%     & -0.89\%     \\
        \bottomrule
    \end{tabular}
    \vspace{-0.25cm}
\end{table}

Such a straightforward fine-tuning approach has been insufficient in classification tasks \cite{liu_fine-pruning_2018}. This could be explained as the orthogonality of model latent space caused by benign and adversarial data \cite{lukas2023pick}. Determining the precise reasons behind fine-tuning's ineffectiveness in removing backdoors remains an area for future investigation.

\paragraph{Identify Backdoors}
Under the assumption that the backdoor objectives are known, we found that triggers can be identified effectively. If such backdoors are detected, the users can simply reject using the backdoored models.

\begin{figure}[htb]
    \centering
    \includegraphics[width=\linewidth]{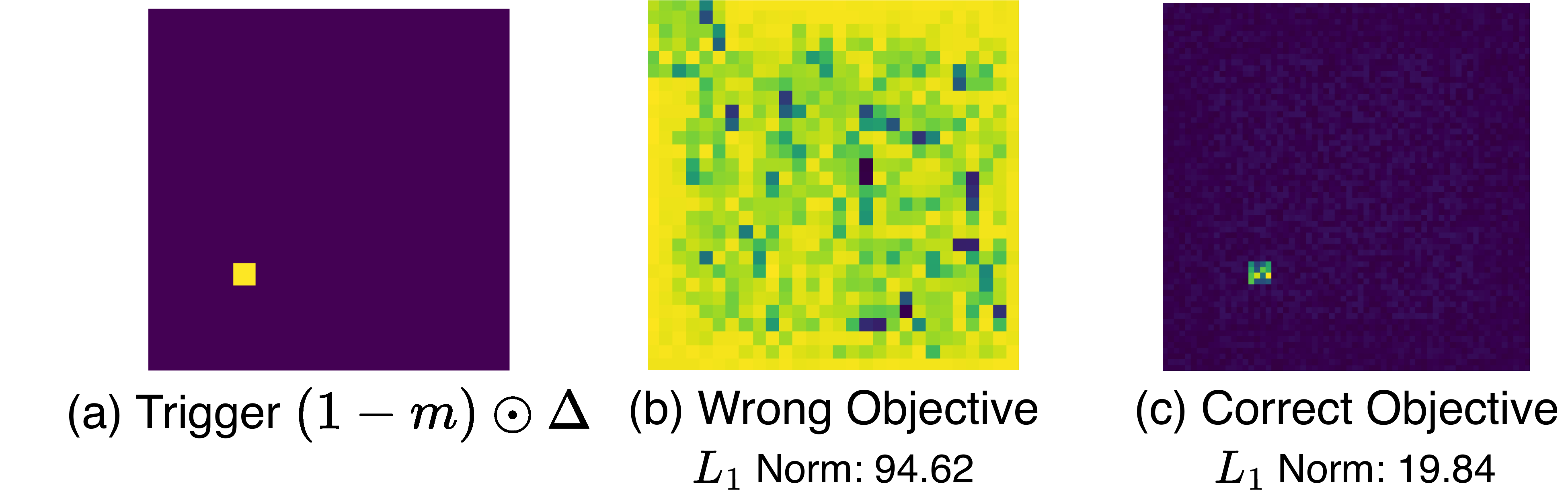}
    \vspace{-0.75cm}
    \caption{\small  Trigger identification on \textsf{Waste-Energy} backdoored neural path planner with wrong (\textsf{Branch}) and correct objectives. }
    \vspace{-0.5cm}
    \label{fig:trigger-inversion}
\end{figure}

An illustration is provided in Fig.~\ref{fig:trigger-inversion}. The trigger inversion technique can find a clear pattern in the reverted trigger images. The results are quantified with average $L_1$ norm pixel-wise between the original trigger and its inverted trigger. Formally, it is defined as:
\begin{align}
    \frac{1}{N} \sum_{i=1}^{N} \frac{1}{H \times W} \sum_{h=1}^{H} \sum_{w=1}^{W} | \mathcal{T}_{i, h, w} - \mathcal{T}'_{i, h, w} |,
    \label{eq:trigger-inversion-metrics}
\end{align}
where $N$ is the number of test maps, $H$ and $W$ are the height and width of the original trigger $\mathcal{T} = (1 - m) \cdot \Delta $, $\mathcal{T}_{i, h, w}$ is the pixel value of the trigger at position $(h, w)$ in the $i$-th test map, and $\mathcal{T}'_{i, h, w}$ is the pixel value of the inverted trigger at position $(h, w)$ in the $i$-th test map.

\begin{table}[!htp]\centering
    \caption{Trigger Inversion Results}\label{tab: trigger-inversion}
    \scriptsize
    \begin{tabular}{l|rrrr|rrrrr}\toprule
            & \multicolumn{4}{c}{Sample-Based} & \multicolumn{4}{c}{Search-Based}                                                                                     \\\cmidrule{2-9}
            & \textbf{TP}                      & \textbf{MD}                      & \textbf{BH} & \textbf{WE} & \textbf{TP} & \textbf{MD} & \textbf{BH} & \textbf{WE} \\\midrule
        DS  & 6.82                             & 4.56                             & 15.19       & 10.24       & 8.15        & 5.13        & 13.20       & 11.92       \\
        PIS & 9.13                             & 6.21                             & 10.21       & 15.20       & 9.19        & 7.09        & 14.99       & 19.84       \\
        \bottomrule
    \end{tabular}
    \begin{tabular}{p{0.98\linewidth}}
        $^1$ TP: \textsf{Trap}, MD: \textsf{Misguide}, BH: \textsf{Branch}, WE: \textsf{Waste Energy} \\
        $^2$ The metric is defined in \eqref{eq:trigger-inversion-metrics}.
    \end{tabular}
    \vspace{-0.25cm}
\end{table}

We demonstrate the results on all the backdoored planners we trained. The results show that the trigger inversion technique can identify the backdoors with a small average $L_1$ norm ($< 19.84$), which shows a clear trigger pattern as demonstrated in Fig.~\ref{fig:trigger-inversion}. However, knowing the objectives is not always possible in practice. We leave the identification of backdoors without knowing the objectives of future work. As an ablation, we also evaluated the trigger inversion technique on the benign model without backdoors, with all the $\phi$ in our experiment. The inverted patterns do not show clear trigger patterns, and are similar to Fig.\ref{fig:trigger-inversion}(b), showing the trigger inversion technique will not report false positives on benign models.

%% file: sections/6_conclusion.tex
\section{Conclusion}

This paper explores the susceptibility of neural path planners to backdoor attacks, highlighting a significant concern in their use within safety-critical domains. Our approach demonstrates how to inject persistent user-specified backdoors into neural planners with high trigger rates and modest performance impact. We also demonstrate potential defenses against our attack and show that simply fine-tuning the neural planner is insufficient to remove backdoors. Trigger inversion, however, can identify backdoors effectively, but with a strong assumption of knowing the planner's objectives. This paper focuses on neural path planning in workspaces, one of the future directions is to extend our work in configuration space, where specifying, injecting, and defending the backdoor behaviors can be challenging due to the high degree of freedom and complex, oftentimes non-differentiable dynamics. We hope this work brings attention to the potential risks of neural path planners and motivates future research on their safety and reliability.